\documentclass[twocolumn, switch]{article} 

\usepackage{preprint}
\usepackage{enumitem}
\usepackage{amsmath, amsthm, amssymb, amsfonts}
\usepackage{algorithm}
\usepackage{algorithmic}

\usepackage[numbers,square]{natbib}

\usepackage[utf8]{inputenc}	
\usepackage[T1]{fontenc}	
\usepackage{xcolor}		
\usepackage[colorlinks = true,
            linkcolor = purple,
            urlcolor  = blue,
            citecolor = cyan,
            anchorcolor = black]{hyperref}	
\usepackage{booktabs} 		
\usepackage{nicefrac}		
\usepackage{microtype}		
\usepackage{lineno}		
\usepackage{float}			

\usepackage{lipsum}		

\usepackage{newfloat}
\DeclareFloatingEnvironment[name={Supplementary Figure}]{suppfigure}
\usepackage{sidecap}
\sidecaptionvpos{figure}{c}

\usepackage{titlesec}
\titlespacing\section{0pt}{12pt plus 3pt minus 3pt}{1pt plus 1pt minus 1pt}
\titlespacing\subsection{0pt}{10pt plus 3pt minus 3pt}{1pt plus 1pt minus 1pt}
\titlespacing\subsubsection{0pt}{8pt plus 3pt minus 3pt}{1pt plus 1pt minus 1pt}

\usepackage{tikz,xcolor,hyperref}

\definecolor{lime}{HTML}{A6CE39}
\DeclareRobustCommand{\orcidicon}{
	\begin{tikzpicture}
	\draw[lime, fill=lime] (0,0)
	circle [radius=0.16]
	node[white] {{\fontfamily{qag}\selectfont \tiny ID}};
	\draw[white, fill=white] (-0.0625,0.095)
	circle [radius=0.007];
	\end{tikzpicture}
	\hspace{-2mm}
}
\foreach \x in {A, ..., Z}{\expandafter\xdef\csname orcid\x\endcsname{\noexpand\href{https://orcid.org/\csname orcidauthor\x\endcsname}
			{\noexpand\orcidicon}}
}

\title{GenTrack: A New Generation of Multi-Object Tracking}

\usepackage{xwatermark}
\newwatermark[firstpage,color=gray!90,angle=0,scale=0.28, xpos=0in,ypos=-5in]{*correspondence: \texttt{toanvn@mmmi.sdu.dk}}

\usepackage{authblk}

\author[1]{Toan Van Nguyen\orcidA{}}
\author[2]{Rasmus G. K. Christiansen\orcidB{}}
\author[3]{Dirk Kraft\orcidC{}}
\author[4]{Leon Bodenhagen\orcidD{}}

\affil[*]{SDU Robotics, University of Southern Denmark}

\begin{document}

\twocolumn[ 
  \begin{@twocolumnfalse} 

\maketitle

\begin{abstract}
This paper introduces a novel multi-object tracking (MOT) method, dubbed GenTrack, whose main contributions include: (1) a hybrid tracking approach employing both stochastic and deterministic manners to robustly handle unknown and time-varying numbers of targets, particularly in maintaining target identity (ID) consistency and managing nonlinear dynamics, (2) leveraging particle swarm optimization (PSO) with some proposed fitness measures to guide stochastic particles toward their target distribution modes, enabling effective tracking even with weak and noisy object detectors, (3) integration of social interactions among targets to enhance PSO-guided particles as well as improve continuous updates of both strong (matched) and weak (unmatched) tracks, thereby reducing ID switches and track loss, especially during occlusions, (4) a GenTrack-based redefined visual MOT baseline incorporating a comprehensive state and observation model based on space consistency, appearance, detection confidence, track penalties, and social scores for systematic and efficient target updates, and (5) the first-ever publicly available source-code reference implementation with minimal dependencies, featuring three variants – GenTrack Simple, Strengthen, and Super – facilitating flexible reimplementation. Experimental results have shown that GenTrack provides superior performance on standard benchmarks and real-world scenarios compared to state-of-the-art trackers, with integrated implementations of baselines for fair comparison. Potential directions for future work are also discussed. The source-code reference implementations of both the proposed method and compared-trackers are provided on GitHub: \href{https://github.com/SDU-VelKoTek/GenTrack}{\url{https://github.com/SDU-VelKoTek/GenTrack}}.   
\end{abstract}

\vspace{0.35cm}

  \end{@twocolumnfalse} 
] 


\section{\textbf{Introduction}}
Multi-object tracking (MOT) aims to accurately and continuously track multiple targets within regions of interest, supporting a wide range of real-world applications. However, achieving robust performance remains challenging due to factors such as irregular motion, occlusions, and visual similarity among objects \cite{ref1,ref2,ref3}. MOT approaches are generally classified into detection-free and detection-based methods \cite{ref2}. Detection-free methods require manual initialization in the first frame and cannot track objects that appear later. In contrast, detection-based methods automatically initialize tracks when objects first enter the scene.

\subsection{Related Works}
Multi-object tracking can be approached through image-based, learning-based or propagation-based methods \cite{ref4, ref5, ref6}. In \cite{ref4}, an image-based tracker, Tracktor, is presented, supporting single-object tracking in noisy conditions and multi-object tracking in uniform settings. In \cite{ref5}, a learning-based method, FairMOT, is built on the anchor-free CenterNet architecture which integrates detection and re-identification through empirical design optimizations, achieving strong performance at the cost of high computational demand and data dependency. In \cite{ref6}, a linear-time tracking-by-propagation model, Samba, is introduced that synchronizes multiple tracklets via autoregressive prediction and long-term memory representations. It effectively handles occlusions without heuristics but exhibits limitations in adapting to abrupt motion changes, reflecting a trade-off between accuracy and memory efficiency.

Due to its ease of use, the Kalman filter is commonly employed to update target motion states and estimate positions, often alongside data association between tracklets and detections. In \cite{ref7}, a pragmatic approach to multi-object tracking (known as SORT) offers an efficient multi-object tracking method for real-time applications, highlighting detection quality as a key performance factor. To reduce identity switches, an extension of SORT is presented in \cite{ref8}, by integrating appearance features, resulting in DeepSORT. In \cite{ref9}, the proposed ByteTrack introduces a simple and effective data association strategy that separates matchings for high and low confidence detections. BoT-SORT in \cite{ref10}, enhances tracking performance by integrating motion and appearance cues, camera-motion compensation, and a refined Kalman filter. In \cite{ref11}, OC-SORT (observation-centric SORT) addresses occlusion-related noise by leveraging object observations to estimate virtual trajectories, reducing error accumulation in parameters of a basic Kalman filter. In \cite{ref12}, an extension of DeepSORT, named StrongSORT, incorporates improvements in detection, feature embedding, and trajectory association, and introduces an appearance-free global association model alongside Gaussian process-based interpolation for handling missed detections. In \cite{ref13}, SMILEtrack is proposed based on ByteTrack and BoT-SORT, in which an object detector is integrated with a Siamese network-based Similarity Learning Module to enhance appearance similarity estimation and matching. In \cite{ref14}, ConfTrack extends a Kalman filter-based tracking-by-detection framework with low-confidence penalization and a cascading algorithm to address noisy detector outputs. Kalman-based trackers typically assume linear motion models and Gaussian noise, focusing primarily on enhancing appearance similarity and matching to improve data association. However, multi-object tracking often involves non-linear dynamics and non-Gaussian noise.

Since the advent of particle filter-based method, significant advancements have been made in single-object tracking. In \cite{ref15}, an unscented particle filter framework is introduced for audio and visual tracking, using the unscented Kalman filter to generate observation-informed proposal distributions. In \cite{ref16}, particle filters combined with multi-mode anisotropic mean shift enable dynamic shape and appearance estimation, along with online reference object learning. The method in \cite{ref17} introduces an incremental likelihood calculation that leverages overlapping regions between particles to improve weight updates. In \cite{ref18}, two high-performance computing schemes are proposed: an alternative Markov Chain Monte Carlo (MCMC) resampling method and a kernel-based approach for particle pipelining. In \cite{ref19}, a particle filter is proposed for visual object tracking with occlusion handling, comprising feature extraction, particle weighting, and occlusion management through a patch-based appearance model utilizing color and motion vector features.  In \cite{ref20}, a particle filter-based tracking algorithm is enhanced by particle swarm optimization to mitigate degeneracy and impoverishment during importance resampling. Correlation filters have also demonstrated advantages in visual object tracking \cite{ref21,ref22}. Building on this, in\cite{ref23}, correlation and particle filters are integrated to leverage their complementary strengths, while a resampling algorithm in \cite{ref24} is leveraged to address degeneracy. In \cite{ref25}, a multi-task correlation particle filter is presented that jointly learns correlation filters by considering interdependencies among multiple features for improved robustness in visual tracking. Furthermore, a chaotic particle filter is introduced in \cite{ref26}, leveraging chaos theory to enhance particle filter performance. Global motion estimation predicts target positions using object dynamics across frames, followed by a color-based particle filter applied within the localized region. In \cite{ref27}, a novel resampling technique based on crow search optimization addresses outlier particles, accelerating convergence in the particle filter tracking framework, alongside an adaptive fusion model integrating multi-cue features per particle. In \cite{ref28}, a mean-shift algorithm is incorporated in a probabilistic filtering framework to manage occlusion, enabling a reduced particle set to preserve multiple modes of the state probability density function. In \cite{ref29}, a particle filtering algorithm is introduced for tracking moving objects in complex real-world environments characterized by shadows, dynamic backgrounds, adverse weather, and illumination changes. A novel scene-dependent feature fusion strategy is proposed, alongside a partitioned template model that captures local object attributes for robust target representation. In \cite{ref30}, an enhanced particle filter integrates quantum particle swarm optimization and an adaptive genetic algorithm to improve particle distribution and sample diversity by resampling, position updating, and fitness-based replacement. In \cite{ref31}, a minimax estimator is combined with sequential Monte Carlo filtering, employing a minimax strategy within a standard particle filter for improved visual tracking.

Notwithstanding prior particle-based approaches, tracking multiple objects presents the challenge of managing the high-dimensional state space that scales with the number of objects. A straightforward approach employs independent single-object particle filters for each target \cite{ref32,ref33}, but this often fails when similar objects interact, causing filter confusion. Multi-object particle filters have been explored \cite{ref34,ref35}, typically assuming a fixed number of targets. In \cite{ref34}, an edge-based multi-object tracking framework is introduced using a variational particle filter, where the proposal distribution derives from an approximated posterior via variational inference instead of the prior in standard Sampling Importance Resampling (SIR). In \cite{ref35}, a discriminative training method is proposed that integrates learning directly into the particle filter inference, optimizing filter parameters based on observed tracking errors. However, in many practical scenarios, targets frequently enter and exit the observation area, necessitating multi-object tracking with a variable number of targets. In \cite{ref36}, a Bayesian filter is proposed for tracking multiple objects when their number is unknown and time-varying, employing a multi-blob likelihood function that assigns comparable likelihoods to hypotheses with differing object counts. By leveraging this approach, a Markov Chain Monte Carlo method \cite{ref37} and a modified Metropolis-Hastings algorithm \cite{ref38} are employed to build a particle filter to address a variable number of interacting targets, as presented in \cite{ref39}, incorporating a Markov random field motion prior to preserve target identities during interactions and reduce tracking failures. In \cite{ref40}, a multi-person tracking-by-detection method, integrated within a particle filtering framework, employs a graded observation model that incorporates continuous confidence scores from pedestrian detectors and online instance-specific classifiers. In \cite{ref41}, an adaptive multi-cue approach is proposed for human tracking, formulating single-target tracking within a Bayesian filtering framework. In \cite{ref42}, a vision system is introduced that combines mixture particle filters with Adaboost to learn, detect, and track objects of interest.

To the best of our knowledge, few studies have explored particle filter-based multi-object tracking with a time-varying number of targets, notably \cite{ref36,ref39,ref40,ref41,ref42}. The approach in \cite{ref36} was among the first to address this problem, but its performance degrades with increasing target count due to reduced accuracy and higher computational cost, limiting its scalability. To mitigate these issues, the method in \cite{ref39} introduced a modified Metropolis-Hastings algorithm incorporating operations such as add-delete and stay-leave to manage dynamic target sets during sampling. While theoretically sound, this approach is prone to errors such as duplicate tracks, especially with limited numbers of particles or iterations. Its effectiveness diminishes as target count increases or fluctuates frequently. Moreover, these works primarily focus on handling variable target numbers, with limited attention to application-oriented aspects such as object scale or appearance variations. In contrast, detection-based approaches \cite{ref40,ref41,ref42} handle target addition and removal more deterministically and emphasize improved observation models, though the optimization of the tracking inference itself is often neglected.

\subsection{Contributions}
This paper presents the philosophy of the proposed GenTrack, and a reformulation of the visual multi-object tracking (MOT) baseline, emphasizing adaptability to diverse tracking tasks. The key contributions include: (1) a hybrid tracking framework leveraging both stochastic and deterministic approaches to handle unknown and time-varying numbers of targets, enhancing robustness and ID consistency under nonlinear dynamics; (2) employing a particle swarm optimization (PSO) algorithm with some proposed fitness measures to guide stochastic particles toward their target distribution modes, enabling effective tracking even with weak and noisy object detectors; (3) integration of social interactions among targets to enhance PSO-guided particles as well as improve continuous updates of both strong (matched) and weak (unmatched) tracks, thereby reducing ID switches and track loss, especially during occlusions. Here, the states of strong tracks are updated by their paired detections while the states of weak tracks are updated by their best fitness particles, velocities, and neighbours; (4) a GenTrack-inspired visual MOT baseline incorporating spatial consistency, appearance, detection confidence, tracking penalties, and social scores into likelihood and fitness functions to support efficient target birth-death and state updates; and (5) a minimal-dependency reference implementation with three variants—GenTrack Simple, GenTrack Strengthen, and GenTrack Super—facilitating reimplementation at different complexity levels. The source-code reference implementation also provides multiple approaches for particle generation and post-PSO resampling options, enabling flexible redevelopment and comparative analysis. Furthermore, implementation insights indicate that overall performance is primarily driven by tracking inference and shows minimal sensitivity to parameter settings. In this paper, fitness and cost measures are formulated to lie within [0, 1] to facilitate systematic interpretability and consistent scaling, simplify integration with other systems or optimization routines. The proposed method is verified through state-of-the-art benchmarks and real-world scenarios, providing superior performance compared to state-of-the-art trackers. Moreover, unlike other particle-based methods, the approach of GenTrack enables it to deliver strong performance with a small number of particles. The proposed GenTrack is designed to operate flexibly across diverse applications, functioning seamlessly with both pre-recorded data and live sensor streams, thereby providing robust support for real-time online tracking. Latency measurements indicate that GenTrack can sustain real-time performance in practical scenarios. Potential directions for future work are also discussed.

\subsection{Organizations}
The remainder of this paper is structured as follows: Section II outlines the methodological background, introduces GenTrack variants, and describes a visual multi-object tracking framework. Section III presents case studies on human and cow tracking. Section IV offers discussions and potential directions for future research, followed by conclusions in Section V.

\section{\textbf{Methodology}}

This section outlines the conceptual foundations of the proposed method, followed by the formulation of GenTrack variants and their application to visual multi-object tracking.

\subsection{Background}
\subsubsection{Hybrid MOT}
Object tracking is typically modelled as a nonlinear system, can be formulated as an instance of Bayesian filtering problem, where the posterior state density is conditionally estimated based on all available measurements, expressed using conditional probability as:

\begin{equation}
\label{equation1}
\begin{split}
& P(X_0, ..., X_t | Z_0, ..., Z_t) = \\ %
& \frac{P(Z_0, ..., Z_t | X_0, ..., X_t). P(X_0, ..., X_t)}{P(Z_0,..., Z_t)}    
\end{split}      
\end{equation}

Here,

\begin{equation}
\label{equation2}
    P(Z_0, ..., Z_t) = P(Z_0). \prod_{i=1}^{t} P(Z_i | Z_{0:i-1})
\end{equation}

\begin{equation}
\label{equation3}
    P(Z_0, ..., Z_t | X_0, ..., X_t) = \prod_{i=0}^{t} P(Z_i | X_i)
\end{equation}

\begin{equation}
\label{equation4}
    P(X_0, ..., X_t) = P(X_0). \prod_{i=1}^{t} P(X_i | X_{i-1})
\end{equation}

A common challenge with this form is the high dimensionality of the state space, which increases over time. To address this, the posterior density $ P(X_0, ..., X_t | Z_0, ..., Z_t) $ can be represented by $ P(X_t | Z_{0:t}) $, and measured as:

\begin{equation}
\label{equation5}
    P(X_t | Z_{0:t}) = \frac{P(Z_t | X_t) . P(X_t | Z_{0:t-1})}{P(Z_t | Z_{0:t-1})}    
\end{equation}

\begin{equation}
\label{equation6}
\begin{split}
   & P(X_t | Z_{0:t-1}) = \\ %
   & \int P(X_t | X_{t-1}).P(X_{t-1} | Z_{0:t-1})\,dX_{t-1}    
\end{split}
\end{equation}

In this context, $P(X_t | X_{t-1})$ is derived from an internal model of the system, while $P(X_{t-1} | Z_{0:t-1})$ is computed at the preceding time step 
$(t-1)$ in accordance with equation (\ref{equation5}). $P(Z_t | X_t)$ corresponds to an observation model of the system. Under the common assumption in many applications that $Z_t$ is conditionally independent on $Z_{0:t-1}$ given $X_t$, the computation of $P(X_t | Z_{0:t})$ can be formulated recursively as follows:

\begin{equation}
\label{equation7}
    \begin{split}
       & P(X_t | Z_{0:t}) \propto \\ %
       & P(Z_t | X_t). \int P(X_t | X_{t-1}) . P(X_{t-1} | Z_{0:t-1})\,dX_{t-1}
    \end{split}
\end{equation}

In a sampling-based approach, the density $P(X_t | Z_{0:t})$ is approximated by a set of \textit{S} weighted random samples $\{X_t^s, W_t^s\}_{s=1}^S$ drawn from it. The observation is then estimated as an unbiased particle approximation, $\{\hat{P} (Z_t | Z_{0:t-1}) = \frac{1}{S}.\sum_{s=1}^{S} P(Z_t | X_{t}^{S})\}$, and the measure of $P(X_t | Z_{0:t})$ can be expressed as:

\begin{equation}
\label{equation8}
    P(X_t | Z_{0:t}) \approx \sum_{s=1}^{S} W_{t}^{s}. \delta{(X_t - X_{t}^{s})}
\end{equation}

Here, $\delta{(\cdot)}$ denotes a Dirac delta function, and $\{X_{t}^{s}\}^{S}$ represents a set of \textit{S} samples generated via the internal model of the system, with corresponding weights measured as follows:

\begin{equation}
\label{equation9}
\begin{split}
    & W_{t}^{s} = \frac{P(Z_t | X_{t}^{s}).P(X_{t}^{s} | Z_{0:t-1})}{P(Z_t | Z_{0:t-1})} \\ %
    & \propto W_{t-1}^{s}. \frac{P(Z_t | X_{t}^{s}).P(X_{t}^{s} | X_{t-1}^{s})}{q(X_{t}^{s} | X_{0:t-1}^{s}, Z_{0:t-1})}
\end{split}
\end{equation}

${q(X_{t}^{s} | X_{0:t-1}^{s}, Z_{0:t-1})}$ is recognized as an important density function, commonly adopted as $P(X_{t}^{s} | X_{t-1}^{s})$, and forms particle weights as: 

\begin{equation}
\label{equation10}
    W_{t}^{s} \propto W_{t-1}^{s}.P(Z_{t} | X_{t}^{s})
\end{equation}

The state is derived via $X_t = \sum_{s=1}^{S} W_{t}^{s}.X_{t}^{s}$, noting $\sum_{s=1}^{S} W_{t}^{s} = 1$. Therefore, normalization $W_{t}^{s} = \frac{W_{t}^{s}}{\sum_{s=1}^{S} W_{t}^{s}}$ is conducted following the measurement of weights in equation (\ref{equation10}). At $t_0$, their initial values are set as $W_{0}^{s} = \frac{1}{S}$.

In systems with a single unique target, the state of the system directly corresponds to the target’s state. For \textit{k} fixed targets, the system can be decomposed into \textit{k} independent subsystems, each representing an individual target, but this approach can lead to multiple particle filters inadvertently tracking the same target under conditions such as occlusion or identical appearance. On the other hand, the system state can be defined as the joint collection of all individual target states $X_t \triangleq \{X_{i,t}\}_{i=1}^{k} $, forming a state space of dimension \textit{kxl}, where \textit{l} is the dimensionality of a single target's state $X_{i,t}$. It is noted that single or fixed-target cases are special instances of more general scenarios where targets frequently enter and leave the observation area, necessitating a multi-object tracker with a variable number of targets. Consequently, maintaining a set of target identifiers within the tracking area is essential. Following \cite{ref36}, the system state is described as $\{K_t, X_t\}$, where $K_t$ represents a set of \textit{k} identifiers of \textit{k} current targets. Henceforth, \textit{k} denotes a time-varying number of targets at the current time step, under the assumption that target entries and leaves are independent of others. By employing Importance Resampling (SIR), the posterior distribution associated with the system state is captured by $\{K_{t}^{s}, X_{t}^{s}, W_{t}^{s}\}_{s=1}^S$ and expressed as follows:

\begin{equation}
\label{equation11}
\begin{split}
    & P(K_t, X_t | Z_{0:t}) \propto \\ %
    & P(Z_t | K_t, X_t). \sum_{S} W_{t-1}^{s}.P(K_{t}^{s}, X_{t}^{s} | K_{t-1}^{s}, X_{t-1}^{s})    
\end{split}
\end{equation}

Here, $P(K_{t}^{s}, X_{t}^{s} | K_{t-1}^{s}, X_{t-1}^{s})$ is estimated using an internal model as follows:

\begin{equation}
\label{equation12}
\begin{split}
    & P(K_t, X_t | K_{t-1}, X_{t-1}) = \\ %
    & P(X_t | K_t, K_{t-1}, X_{t-1}).P(K_t | K_{t-1}, X_{t-1})    
\end{split}
\end{equation}

Equation (\ref{equation11}) involves a sampling process $\{K_{t}^{S}, X_{t}^{S}\} \sim q(K_t, X_t) \triangleq \sum_{s} W_{t-1}^{s}.P(K_t, X_t | K_{t-1}^{s},X_{t-1}^{s})$ that meets well-known limitations such as inefficiency in high-dimensional spaces, weight degeneracy, sensitive the proposal distribution, and high computational cost. To address these issues, a Markov Chain Monte Carlo (MCMC) approach can be employed, reformatting the posterior density as follows:

\begin{equation}
\label{equation13}
\begin{split}
    & P(K_t, X_t | Z_{0:t}) \propto \\ %
    & P(Z_t | K_t, X_t). \sum_{S} P(K_{t}^{s}, X_{t}^{s} | K_{t-1}^{s}, X_{t-1}^{s})       
\end{split} 
\end{equation}

The sampling is now defined as $\{K_{t}^{S}, X_{t}^{S}\} \triangleq \sum_{s} P(K_t, X_t | K_{t-1}^{s},X_{t-1}^{s})$ but challenges remain due to the complex internal model in equation (\ref{equation12}), which not only predicts target motion but also accounts for targets entering and leaving the tracking area. This issue was addressed in \cite{ref39} by partitioning state $(K_t, X_t)$ into entering $(K_E, X_E)$ and staying $(K_S, X_S)$, then using a Reversible Jump MCMC particle filter \cite{ref37} with add-delete and stay-leave moves, and a modified Metropolis-Hastings algorithm \cite{ref38} to handle variable target numbers. Although this approach resolves the variable target count by dynamically creating and removing tracks during sampling, it incurs high computational costs and risks track duplication, especially with few particles, reducing performance when target numbers change frequently. 

Stochastic approaches handle uncertainty and nonlinear systems better than deterministic ones but yield variable outputs across runs due to the randomness in particle generation, especially in detection-free tracking. Even detection-based methods that incorporate advanced single-object trackers to refine bounding boxes will also lead to inconsistent results between executions, which is problematic for multi-object tracking requiring consistent IDs. Conversely, deterministic methods provide consistent outputs, such as tracking-by-detection with data association, though their performance depends on detection and motion model quality. This paper presents a hybrid multi-object tracking framework, using the above stochastic approach for handling non-linear, non-Gaussian motions, while a deterministic manner is employed for consistent target identifiers. Although the deterministic component requires an object detector for data association, its limitations are mitigated by the stochastic counterpart, enabling robust tracking despite weak detectors. As a result, the proposed hybrid MOT effectively handles complex system dynamics, maintains global output consistency, and preserves target identities despite frequent changes in target numbers. Here, the sampling is conducted in the manner described by equation (\ref{equation13}) but does not directly create or remove tracks during sampling. Instead, each target is defined with a unique birth and a unique death, where birth corresponds to track initiation and death to track removal. The birth-death is determined by the history of associations between tracklets and detections, represented via $P(K_t, X_t | Z_t, K_{t-1}^{h}, X_{t-1}^{h}, K_{t-1}, X_{t-1})$. The associations are evaluated using a cost, as follows:

\begin{equation}
\label{equation14}
    C_{\varepsilon} \xleftarrow{} P(K_t, X_t | Z_t, K_{t-1}^{h}, X_{t-1}^{h}, K_{t-1}, X_{t-1})
\end{equation}

Here, $\{K_{t}^{h}, X_{t}^{h}\}$ denotes the tracking history of $\{K_{t}, X_{t}\}$, reflecting track confidence over time. It should be noted that the current detections $\{D_t\}$ and their associated confidences $\{D_{t}^{conf}\}$ are also incorporated into $Z_t$. A detailed presentation of the observations will be provided in the section on GenTrack variants. 

\vspace{11pt}
\subsubsection{PSO-Guided Particles}
Sampling typically begins with the system internal model, expressed as $\{K_{t}^{S}, X_{t}^{S}\} \triangleq \sum_{s} P(K_t, X_t | K_{t-1}^{s},X_{t-1}^{s})$. However, object motion is unpredictable, and its motion model is normally unavailable in multi-object tracking. Thus, a random motion model is employed. Consequently, particles are inherently stochastic and require guidance toward optimal locations to enable accurate data association and convergence. It is noted that the hybrid MOT framework leverages strengths of stochastic and deterministic approaches to deal with uncertainty and non-linear system while ensuring consistent tracking outputs across runs. Target birth-death resulted from deterministic data association and excluded from particle sampling. Full posterior sampling is unnecessary, instead, the focus now is on generating optimal particles per target to facilitate convergence toward their respective target distribution modes. By virtue of its rapid convergence and efficient exploration of local likelihood modes, PSO \cite{ref43} is therefore well-suited for the proposed hybrid MOT framework, particularly in scenarios with numerous targets requiring real-time tracking. This paper introduces tailored PSO fitness functions $f_{PSO}$, adapted to different variants of GenTrack.

\begin{algorithm}[H]
\caption{Pseudocode of PSO Integrated in The Proposed Hybrid MOT.}\label{alg:alg1}
\begin{algorithmic}
\STATE 
\STATE {\textsc{Initialize Particle Sets}} $\{K_{PSO}^{s}, X_{PSO}^{s}, V_{PSO}^{s}\}_{s=1}^{S}$
\STATE {\textsc{While}} {(! termination condition)}:
\STATE \hspace{0.3cm} \textbf{For \textit{s} = 1 to \textit{S}}: 
\STATE \hspace{0.6cm} \textbf{If} $f(K_{p}^{s}, X_{p}^{s}) < f(K_{PSO}^{s}, X_{PSO}^{s})$:
\STATE \hspace{0.9cm} $(K_{p}^{s}, X_{p}^{s}, V_{p}^{s}) \xleftarrow{} (K_{PSO}^{s}, X_{PSO}^{s}, V_{PSO}^{s})$
\STATE \hspace{0.6cm} \textbf{If} $f(K_{g}^{s}, X_{g}^{s}) < f(K_{p}^{s}, X_{p}^{s})$:
\STATE \hspace{0.9cm} $(K_{g}^{s}, X_{g}^{s}, V_{g}^{s}) \xleftarrow{} (K_{p}^{s}, X_{p}^{s}, V_{p}^{s})$
\STATE \hspace{0.6cm} \textbf{For \textit{j} = 1 to \textit{l}}: 
\STATE \hspace{0.9cm} $r_p = U(0,1), r_g = U(0,1)$
\STATE \hspace{0.9cm} ${}^jV_{PSO}^{s,max} \xleftarrow{} (K_{t-1}, X_{t-1}, V_{t-1})$
\STATE \hspace{0.9cm} ${}^jV_{PSO}^{s} = \eta.{}^jV_{PSO}^{s} + r_p.\varphi_p. ({}^jX_{p}^{s}-{}^jX_{PSO}^{s})$ \\ \hspace{4.45cm} $+ r_g.\varphi_g.({}^jX_{g}^{s} - {}^jX_{PSO}^{s})$ 
\STATE \hspace{0.9cm} ${}^jX_{PSO}^{s} += sign({}^jV_{PSO}^{s}).min(|{}^jV_{PSO}^{s}|,{}^jV_{PSO}^{max})$ 
\STATE {\textsc{Return}} $\{K_{p}^{s}, X_{p}^{s}, V_{p}^{s}\}_{s=1}^{S}$ and $\{K_{g}^{s}, X_{g}^{s}\}$

\vspace{11pt}
\STATE {\textsc{Notes:}} $U(a,b)$ denotes a uniform random on $(a, b)$, and \textit{l} denotes the state space of a single target. For convergence \cite{ref44}, parameters satisfy $\eta \in (0,1), \varphi_p, \varphi_g \in (1,3)$.$(K_{p}^{s}, X_{p}^{s})$ and $(K_{g}^{s}, X_{g}^{s})$ denote the personal and global bests in PSO iterations. Particles with low fitness may be discarded or replaced by global bests to steer sampling toward the mode of the target distribution. Both approaches are included in the source code for comparison:
\vspace{2pt}
\STATE \textbf{For \textit{s} = 1 to \textit{S}} 
\STATE \hspace{0.5cm} \textbf{If} $f(K^{s}, X^{s}) < \sigma_0$
\STATE \hspace{1.0 cm} Discard $(K^{s}, X^{s})$, or $(K^s, X^s) = (K_g^s, X_g^s)$
\end{algorithmic}
\label{alg1}
\end{algorithm}

Algorithm 1 presents the PSO pseudocode for the proposed hybrid MOT, where $V_t$ denotes velocity. Initial particle sets $\{K_{PSO}^{s}, X_{PSO}^{s}, V_{PSO}^{s}\}_{s=1}^{S}$ are generated using a random motion model, either from previous particle sets $\{K_{t-1}^{s}, X_{t-1}^{s}, V_{t-1}^{s}\}_{s=1}^{S}$ or the previous optimal target state $\{K_{t-1}, X_{t-1}, V_{t-1}\}$, with added perturbations $U_X$ and $U_V$. Here, $U_X$ and $U_V$ represent random position and velocity changes, bounded by $U_{X}^{max}$ and $U_{V}^{max}$, derived from $\{U_{X}^{max},U_{V}^{max}\} \xleftarrow{} \{K_{t-1}, X_{t-1}, V_{t-1}\}$. The resulting forms are either $\{K_{PSO}^{s}, X_{PSO}^{s}, V_{PSO}^{s}\} \xleftarrow{} \{K_{t-1}^{s}, X_{t-1}^{s}, V_{t-1}^{s},U_X, U_V\}$ or $\{K_{PSO}^{s}, X_{PSO}^{s}, V_{PSO}^{s}\} \xleftarrow{} \{K_{t-1}, X_{t-1}, V_{t-1}, U_X, U_V\}$, respectively.

The fitness function is formulated to balance between consistency and exploration. Here, consistency refers to the alignment between a particle and the previous optimal state of its target, whereas exploration pertains to the particle deviation from its own update in the preceding PSO iteration, expressed as:

\begin{equation}
\label{equation15}
    f_{PSO} \xleftarrow{} (f_{PSO}^{h}, f_{PSO}^{p})
\end{equation}

Here, $f_{PSO}^{h} = f(\{K_{PSO}^{s}, X_{PSO}^{s}\}, \{K_{t-1}, X_{t-1}\})$ denotes the historical fitness between the current particle and the previous optimal state of its target, while $f_{PSO}^{p} = f(\{K_{PSO}^{s}, X_{PSO}^{s}\}, \{K_{PSO}^{s'}, X_{PSO}^{s'}\})$ represents the exploratory fitness comparing the current particle to its own update in the previous PSO iteration. Both metrics are derived from the same function $f(\bullet, \bullet)$, which is a defined according to the specific tracking application and will be detailed in later sections.

A common challenge in multi-object tracking is the frequent proximity of targets, which introduces interactions that impact tracking accuracy. Increased neighbouring objects lead to greater noise and a higher possibility of identity switches. Thus, it is essential to incorporate interactions between particles and neighbours of their targets, thereby refining the fitness measure as follows:

\begin{equation}
\label{equation16}
    f_{PSO} \xleftarrow{} (f_{PSO}^{h}, f_{PSO}^{p}, f_{PSO}^{i,s})
\end{equation}

Here, $f_{PSO}^{i,s} = f(\{K_{PSO}^{s}, X_{PSO}^{s}\},\{K_{t}^{n}, X_{t}^{n}\}_{n=1}^{N})$ represents the social fitness between the current particle and N neighbours of its target.

\vspace{11pt}
\subsubsection{Social Contributions}
The interactions among targets in multi-object tracking are unavoidable and significantly impact tracking performance, particularly under occlusion. These interactions can result in ID switches or track losses, especially when a target is heavily occluded. An increased number of neighbouring objects introduces more noise and raises the likelihood of ID switches, in which the influence of one object on another’s tracking performance depends on specific criteria. In this context, the neighbours of object $(K_{i,t}, X_{i,t}) \in (K_t, X_t)$ are denoted as $\{K_{i,t}, X_{i,t}\}^{nei} = \Psi \langle (K_{i,t}, X_{i,t}), \varepsilon_{nei} \rangle$, and $\Psi \langle \hat{\ },\hat{\ } \rangle$ performs a nearest-neighbour search to identify objects near $(K_{i,t}, X_{i,t})$ within a threshold $\varepsilon_{nei}$. The threshold $\varepsilon_{nei}$ is determined by a state-based adaptive neighbour range search, which is dynamically adjusted for each $X_{i,t}$. Incorporating target interactions \cite{ref39}, equation (\ref{equation13}) is reformulated as follows:

\begin{equation}
\label{equation17}
\begin{split}
    & P(K_t, X_t | Z_{0:t}) \propto P(Z_t | K_t, X_t). \\ %
    & \sum_{S} P(K_{t}^{s}, X_{t}^{s} | K_{t-1}^{s}, X_{t-1}^{s}).\prod_{g,q}^{N}\Omega(X_{g,t},X_{q,t})       
\end{split} 
\end{equation}

Here, $\Omega(X_{g,t},X_{q,t})$ demonstrates the dynamic interaction between $(K_{g,t}, X_{g,t})$ and $(K_{q,t}, X_{q,t})$ and \textit{N} denotes the number of the neighbouring ensemble.

It is noted that the birth and death processes of targets are determined by the historical evolution of deterministic data associations, as formalized in equation (\ref{equation14}). Although the system is described by a joint state, the sampling procedure emphasizes the local states of individual targets, as outlined in the preceding section. Accordingly, interactions may be considered at the individual level, whereby an instance of equation (\ref{equation17}) can be formulated to characterize the behaviour of a single entity within the system, as: 

\begin{equation}
\label{equation18}
\begin{split}
    & P(K_{i,t}, X_{i,t} | Z_{0:t}) \propto P(Z_t | K_{i,t}, X_{i,t}). \\ %
    & \sum_{S} P(K_{i,t}^{s}, X_{i,t}^{s} | K_{i,t-1}^{s}, X_{i,t-1}^{s}).\prod_{n}^{N}\Omega(X_{i,t},X_{j,t}^{n})       
\end{split} 
\end{equation}

Henceforth, \textit{N} denotes the number of neighbours $(K_{j,t}, X_{j,t})$ associated with $(K_{i,t}, X_{i,t})$, defined as $\{K_{i,t}, X_{i,t}\}^{nei} = \{K_{j,t}^{n}, X_{j,t}^{n}\}_{n=1}^{N}$.

The interactions among objects will contribute to PSO-guided particle behaviour and state updates within the GenTrack Super framework. These interactions do not directly apply to data association between tracklets and detections, as this could introduce ambiguity and false positives. Instead, they are designed to guide target particles diverge from neighbouring states, thereby preventing direct convergence onto neighbouring target states, which is particularly beneficial in occlusion scenarios. The social fitness of the particle associated with the $i^{th}$ target is defined as:

\begin{equation}
\label{equation19}
    f_{PSO}^{i,s} \xleftarrow{} (\{K_{i,t}^{s}, X_{i,t}^{s}\}, \{K_{i,t},X_{i,t}\}^{nei})
\end{equation}

By incorporating social contributions, the state of the $i^{th}$ target is updated as:

\begin{equation}
\label{equation20}
\begin{split}
     (K_{i,t}, X_{i,t}) \xleftarrow{} (K_{i,t-1}^{h}, X_{i,t-1}^{h}, K_{i,t-1},X_{i,t-1}, \\ %
     \{K_{i,t},X_{i,t}\}^{nei})    
\end{split}
\end{equation}

\subsection{GenTrack Variants}
The GenTrack framework, outlined in Algorithm 2, encompasses three variants – GenTrack Simple, GenTrack Strengthen, and GenTrack Super – distinguished by their particle sampling strategies and state update mechanisms. Target tracking history is denoted as $\{K_{t}^{h}, X_{t}^{h}\} = \{K_t, X_{t}^{pen}, X_{t}^{age}\}$, where $X_{t}^{pen}$ and $X_{t}^{age}$ represent track penalties and ages of targets, respectively, based on the history of target tracks up to the current frame $I_t$. The deterministic data association produces $\{K_{i,t}^{m}, X_{i,t}^{m}, det_{j,t}^{m}\}_{m=1}^{M}$, a set of \textit{M} pairs, each consisting of a track and a current detection, $\{K_t^m,X_t^m\} \in \{K_t, X_t\}$, $\{det_t^m\} \in \{D_t\}$.

\begin{algorithm}[H]
\caption{GenTrack Framework.}\label{alg:alg2}
\begin{algorithmic}
\STATE 
\STATE {\textsc{Inputs:}} $\{K_{t-1}, X_{t-1}, V_{t-1}, X_{t-1}^{pen}, X_{t-1}^{age}, I_t,D_t,D_t^{conf}\}$ 
\STATE {\textsc{Particle Sampling}}:
\STATE \hspace{0.5cm} \textsc{Return} $\{K_t^s, X_t^s\}_{s=1}^S$ 
\STATE {\textsc{Deterministic Data Association}} $(\{K_t^s,X_t^s\}_{s=1}^S, $
\STATE $X_{t-1}^{pen},X_{t-1}^{age},D_t,D_t^{conf})$:
\STATE \hspace{0.5cm} \textsc{Return} $\{K_{i,t}^m, X_{i,t}^m,det_{j,t}^m\}_{m=1}^M$ 
\STATE {\textsc{State Updates}}
\STATE {\textsc{Outputs:}} $\{K_t,X_t,V_t,X_t^{pen},X_t^{age}\}$
\end{algorithmic}
\label{alg2}
\end{algorithm}

This paper employs the Hungarian algorithm \cite{ref45} for data association. It is noted that Hungarian Assignment can be applied by merging all target particles into a list and matching them to detections. Assigning a unique detection to a target by selecting it from the best particle-detection pair among its associated pairs ensures one-to-one matching but may result in particles from the same target being matched to different detections. This can lead to more weak tracks and unmatched detections, as well as increased computational costs. To overcome this, a target-oriented cost matrix $C_{mat} \in R^{TxD}$ is proposed, where \textit{T} and \textit{D} denote the numbers of current targets and detections, respectively. The cost is measured as:

\begin{equation}
\label{equation21}
\begin{split}
     C_{i,j} = \lambda_p.\frac{1}{S}. \sum_{s=1}^S C_m^{X_{i,t}^s,det_j}  + \lambda_d.(1-det_j^{conf}) \\ %
     + \lambda_h.X_{i,t-1}^{pen}
\end{split}
\end{equation}

Here, the matching cost between $X_{i,t}$ and $det_j$ is represented by $C_{i,j}$. \textit{S} denotes the number of particles of $X_{i,t}$. $C_m^{X_{i,t}^s,det_j} \in [0, 1]$ denotes the motion cost between $X_{i,t}^s$ and $det_j$. The previous state penalty $X_{i,t-1}^{pen} \in [0, 1]$ is derived from its matching history, while $det_j^{conf} \in [0, 1]$ indicates the confidence of detection $det_j$. The positive values $\lambda_h, \lambda_d$, and $\lambda_p$ control the contribution of track penalty, detection confidence, and particles, here $\lambda_h + \lambda_d + \lambda_p =1$.

Based on $C_{mat}$, the pairs $\{K_{i,t}^{m}, X_{i,t}^{m}, det_{j,t}^{m}\}_{m=1}^{M}$ are obtained, where $(K_{i,t}^m,X_{i,t}^m)$ constitutes a strong track. The current weak tracks are then defined as the residual set: $\{K_t^w,X_t^w\} = \{K_t,X_t\} - \{K_t^m, X_t^m\}$. Unmatched detections are given by $\{det_t^u,det_t^{u,conf}\} = \{D_t, D_t^{conf}\} - \{det_t^m,det_t^{m,conf}\}$, and are used to initialize new tracks $\{K_t^n, X_t^n\}$. The overall system state is thus defined as: $\{K_t,X_t\} = \{(K_t^m,K_t^w,K_t^n),(X_t^m,X_t^w,X_t^n)\}$. For all three GenTrack variants, the states of $(K_t^m,X_t^m)$ and $(K_t^n,X_t^n)$ are subsequently updated as:   

\begin{equation}
\label{equation22}
\begin{cases}
    \{K_t^m,X_t^m\} \xleftarrow{} \{det_t^m\} \\
    \{K_t^n,X_t^n\} \xleftarrow{} \{det_t^u,det_t^{u,conf}\} \\
    X_t^{m,pen} = 0 \\
    X_t^{n,pen} = 0 \\
    X_t^{m,age} = 0 \\
    X_t^{n,age} = 0 \\
\end{cases}
\end{equation}

\subsubsection{GenTrack Simple}
In GenTrack Simple, particle sampling for targets is performed using a random motion model. Each particle state $\{K_t^s,X_t^s,V_t^s\}_{s=1}^S$ is generated either from its corresponding prior state $\{K_{t-1}^s,X_{t-1}^s,V_{t-1}^s\}_{s=1}^S$ as $\{K_t^s,X_t^s,V_t^s\} \xleftarrow{} \{K_{t-1}^s,X_{t-1}^s,V_{t-1}^s,U_X,U_V\}$, or from the previous optimal state of target $\{K_{t-1},X_{t-1},V_{t-1}\}$ as $\{K_t^s,X_t^s,V_t^s\} \xleftarrow{} \{K_{t-1},X_{t-1},V_{t-1},U_X,U_V\}$. Here, $U_X,U_V$ are random perturbations on position and velocity, bounded by $U_X^{max}$ and $U_V^{max}$, with $\{U_X^{max},U_V^{max}\} \xleftarrow{} \{K_{t-1},X_{t-1},V_{t-1}\}$. Besides, the state of weak tracks $(K_t^w,X_t^w)$ are updated based on their own motion dynamics:    

\begin{equation}
\label{equation23}
\begin{cases}
    \{K_t^w,X_t^w\} \xleftarrow{} \{K_{t-1}^w,X_{t-1}^w,V_{t}^w\} \\
    X_t^{w,pen} = X_{t-1}^{w,pen} + \frac{\rho_{max}}{\partial_{max}} \\
    X_t^{w,age} = X_{t-1}^{w,age} + 1\\
\end{cases}
\end{equation}

Here, $\rho_{max}$ and $\partial_{max}$ denote the maximum penalty and maximum age, respectively. As defined in equation(\ref{equation21}), to adhere to the probability constraints, $\rho_{max} = 1$.

\vspace{11pt}
\subsubsection{GenTrack Strengthen}

In GenTrack Strengthen, the initial particle sets are derived from the motion model, as in GenTrack Simple. These particles are then refined using a PSO procedure guided by the fitness function defined in equation (\ref{equation15}). The optimization yields both the refined particle sets of $\{K_t^s,X_t^s\}_{s=1}^S$ and the global best estimates for all targets, as $\{K_g^i, X_g^i,f_{g,PSO}^{i,h}\}_{i=1}^k$. The refined particle sets serve as inputs of Hungarian algorithm, while the global best estimates are employed to update the state of weak tracks $(K_t^w,X_t^w)$ as:

\begin{equation}
\label{equation24}
\begin{cases}
    \{K_t^w,X_t^w\} \xleftarrow{} \{K_{g}^w,X_{g}^w,f_{g,PSO}^{w,h}, V_t^w\} \\
    X_t^{w,pen} = X_{t-1}^{w,pen} + sign(\rho_{re}-f_{g,PSO}^{w,h}).\Delta_{pen} \\
    X_t^{w,age} = X_{t-1}^{w,age} + sign(\rho_{re}-f_{g,PSO}^{w,h})\\
\end{cases}
\end{equation}

Here, $\rho_{re} \in [0, 1]$ denotes the track recovery threshold. The penalty increment is defined as $\Delta_{pen} = \frac{\rho_{max} + C_f}{\partial_{max}} = \frac{\rho_{max}.(2-f_{g,PSO}^{w,h})}{\partial_{max}}$, where $C_f = \rho_{max}.(1-f_{g,PSO}^{w,h})$. Track penalties and ages are constrained by: $\rho_{max} \ge X_t^{w,pen} \ge 0, X_t^{w,age} \ge 0$.    

\vspace{11pt}
\subsubsection{GenTrack Super}

In GenTrack Super, the initial particle sets are generated using the motion model, as GenTrack Simple. A PSO procedure is then applied to refine these particles based on the fitness functions defined in equations (\ref{equation16}) and (\ref{equation19}). In addition to the optimized particle sets of $\{K_t^s, X_t^s\}_{s=1}^S$, the PSO returns global bests and neighbours of all targets $\{K_g^i, X_g^i,f_{g,PSO}^{i,h}, X_i^{nei}\}_{i=1}^k$. Here, the optimized particle sets serve as inputs of Hungarian algorithm, while $\{K_g^i, X_g^i,f_{g,PSO}^{i,h}, X_i^{nei}\}_{i=1}^k$ is subsequently used to update the state of weak tracks $(K_t^w,X_t^w)$ via a modified form of equation (\ref{equation20}) incorporating social PSO dynamics: 

\begin{equation}
\label{equation25}
\begin{cases}
    \{K_t^w,X_t^w\} \xleftarrow{} \{K_{g}^w,X_{g}^w,f_{g,PSO}^{w,h}, V_t^w, X_w^{nei}\} \\
    X_t^{w,pen} = X_{t-1}^{w,pen} + sign(\rho_{re}-f_{g,PSO}^{w,h}).\Delta_{pen} \\
    X_t^{w,age} = X_{t-1}^{w,age} + sign(\rho_{re}-f_{g,PSO}^{w,h})\\
\end{cases}
\end{equation}

Although the two final lines in equations (\ref{equation24}) and (\ref{equation25}) are identical, the social contribution is already incorporated in $f_{g,PSO}^{w,h}$ of equation (\ref{equation25}), as it is obtained through PSO using equations (\ref{equation16}) and (\ref{equation19}). Moreover, if a weak track has no neighbours after PSO, an expanded search $\Psi \langle (K_{i,t},X_{i,t}), \varepsilon_{nei}^{exp} \rangle$ is performed, where $\varepsilon_{nei}^{exp}$ represents an expanded neighbour condition. Furthermore, if the neighbour is a strong track, its state is determined by its matched current detection. Otherwise, if the neighbour is a weak track, its state corresponds to its previous optimal state.

For all GenTrack variants, track penalties and ages are updated as $\{X_t^{pen}\} = \{X_t^{m,pen}, X_t^{w,pen}, X_t^{n,pen}\}$ and $\{X_t^{age}\} = \{X_t^{m,age},X_t^{w,age},X_t^{n,age}\}$, respectively. Finally, expired tracks and their associated information are subsequently removed from the tracklets: 

\begin{equation}
\label{equation26}
\begin{split}
    & \{K_t,X_t,V_t,X_t^{pen},X_t^{age}\} = \\ %
    & \{K_{i,t},X_{i,t},V_{i,t},X_{i,t}^{pen},X_{i,t}^{age} | X_{i,t}^{age} < \partial_{max}\}_{i=1}^k
\end{split}
\end{equation}

It is noted that the velocity of target will be smoothened for the next frame as follows: $V_t = \frac{V_t +V_{t-1}}{2}$. Besides, to enhance application performance, dynamic and entrance penalties are incorporated into all GenTrack variants to account for track age and penalty, thereby expediting the removal of expired tracks. These factors are implemented in the source, with entrance areas are defined as the image borders.

\subsection{Visual Multi-Object Tracking}
\subsubsection{Object State}

To account for temporal variations in object appearance, the state of an object at time \textit{t} is defined as $X_{i,t} = (u_{i,t},v_{i,t},w_{i,t},h_{i,t})$, where $(u_{i,t},v_{i,t})$ denotes the bounding box center and $(w_{i,t},h_{i,t})$ denotes its width and height in image pixel coordinates. The corresponding velocity is $V_{i,t} = (\dot{u}_{i,t}, \dot{v}_{i,t}, \dot{w}_{i,t}, \dot{h}_{i,t})$. The multi-object state is given by $\{K_t,X_t\} = \{K_{i,t}, X_{i,t}\}_{i=1}^k$. To support GenTrack variants, the tracklets are extended to include velocity, penalty, and age terms, as: $\{K_t,X_t,V_t,X_t^{pen},X_t^{age}\} = \{K_{i,t},X_{i,t},V_{i,t},X_{i,t}^{pen},X_{i,t}^{age}\}_{i=1}^k$.  

\vspace{11pt}
\subsubsection{Internal Model}

In multi-object tracking, object motions are inherently unpredictable, and explicit motion models are typically unavailable, particularly for natural entities such as humans or animals. Consequently, this paper employs a random motion model defined as $\{K_t,X_t,V_t\} \xleftarrow{} \{K_{t-1},X_{t-1},V_{t-1},U_X,U_V\}$, where, $U_X$ and $U_V$ represent stochastic perturbations in position and velocity. The velocities of targets in the system $V_t = \{V_{1,t}, ..., V_{k,t}\}$ are constrained within bounds $V_t^{max} = \{V_{1,t}^{max}, ..., V_{k,t}^{max}\}$, such that $V_{i,t} \in (-V_{i,t}^{max}, V_{i,t}^{max})$, with $V_{i,t}^{max} = (\dot{u}_{i,t}^{max}, \dot{v}_{i,t}^{max}, \dot{w}_{i,t}^{max}, \dot{h}_{i,t}^{max})$. Similarly, the random position and velocity changes $U_{X_{i,t}} = (U_{i,t}^u, U_{i,t}^v, U_{i,t}^w, U_{i,t}^h)$ and $U_{V_{i,t}} = (U_{i,t}^{\dot{u}}, U_{i,t}^{\dot{v}}, U_{i,t}^{\dot{w}}, U_{i,t}^{\dot{h}})$ are bounded by $(-U_{X_{i,t}}^{max}, U_{X_{i,t}}^{max})$ and $(-U_{V_{i,t}}^{max}, U_{V_{i,t}}^{max})$, respectively, with bounds dependent on the prior state $X_{i, t-1}$, facilitating adaptive behavior across different targets. The motion model can be succinctly described as:    

\begin{equation}
\label{equation27}
\begin{cases}
    V_{i,t} = V_{i,t-1} + \varepsilon_V.U_{V_{i,t}} \\
    X_{i,t} = X_{i,t-1} + \lambda_V.V_{i,t} + \lambda_X.\varepsilon_X.U_{X_{i,t}} \\
\end{cases}
\end{equation}

Here, $\varepsilon_X, \varepsilon_V, \lambda_X$, and $\lambda_V$ denote state exploration, velocity exploration, state contribution, and velocity contribution, respectively. 

\vspace{11pt}
\subsubsection{Observation Model}

The observation model in visual multi-object tracking commonly incorporates appearance similarity and motion consistency, measured by the intersection of union (IoU) between predicted (typically using a Kalman filter) and detected bounding boxes. This paper proposes extended observation models tailored to the proposed GenTrack variants. First, the full PSO fitness function in equation (\ref{equation16}) comprises three components: $f_{PSO}^h, f_{PSO}^p$ and $f_{PSO}^i$, corresponding to history fitness (between the current particle and the previous optimal state of its target), exploration fitness (between the current particle and its update from the prior PSO iteration), and social fitness (between the particle and the neighbours of its target), respectively. Notably, $f_{PSO}^h$ and $f_{PSO}^p$ are measured using the same function $f(\bullet, \bullet)$, differing only in their inputs according to the fitness type. 

\begin{equation}
\label{equation28}
    f(\bullet,\bullet) = \lambda_s.f_s + \lambda_m.f_m
\end{equation}

The appearance similarity fitness $f_s$ is defined as the cosine similarity between Histogram of Oriented Gradients (HoG) feature vectors $\vec{X}_i^s$ and $\vec{X}_j$, given by $f_s = \frac{\langle \vec{X}_i^s \bullet \vec{X}_j \rangle}{|\vec{X}_i^s|.|\vec{X}_j|}$, with $\vec{X}_i^s$ and $\vec{X}_j$ are extracted from image areas of $X_i^s$ and $X_j$, where $X_i^s = (u_i^s, v_i^s, w_i^s, h_i^s)$ and $X_j = (u_j, v_j, w_j, h_j)$ denote bounding boxes, are inputs of $f(\bullet,\bullet)$. The motion fitness $f_m$ is expressed as $f_m = 1- \frac{min(|X_i^s -X_j|,d_{o,m})}{d_{o,m}}$, where $d_{o,m}$ represents the maximum allowed distance between $X_i^s$ and $X_j$, obtained from $(w_i, h_i)$ and $(w_j, h_j)$. The positive values $ \lambda_s$ and $\lambda_m$ control the contribution of appearance similarity and motion, here $\lambda_s + \lambda_m = 1$. Besides, $|x|$ denotes the magnitude of a vector \textit{x}, and $\langle a \bullet b \rangle$ denotes the dot product of vectors $\vec{a}$ and $\vec{b}$. It is noted that the HoG feature vectors are designed as non-negative, thus $f_s \in [0, 1]$.

The metric for social fitness $f_{PSO}^{i,s}$ will be assessed as follows:

\begin{equation}
\label{equation29}
\begin{split}
     &f_{PSO}^{i,s} = \frac{\xi_p}{N}.\sum_{j=1}^N \frac{min(|X_{i,t}^s - X_{j,t}|, 2\varepsilon_{nei})}{2\varepsilon_{nei}} \\ %
     &+ \frac{\xi_V}{N}.\sum_{j=1}^N \frac{min(|V_{i,t}^s - V_{j,t}|, V_s^{max})}{V_s^{max}}
\end{split}
\end{equation}

Here, $\varepsilon_{nei}$ denotes the state-based adaptive neighbour range search for each $X_{i,t}$, defined as the diagonal of its bounding box. \textit{N} represents the number of neighbours $X_j$ of $X_i$, and $V_s^{max} = V_{i,t}^{max} + U_{V_{i,t}}^{max}$. The positives $\xi_p$ and $\xi_V$ control contributions of state and velocity of particle to its social fitness, $\xi_p + \xi_V = 1$. When $N=0, f_{PSO}^{i,s} = 1$.

An instance of equation (\ref{equation15}) is then given by $f_{PSO} = \sigma_h.f_{PSO}^h + \sigma_p.f_{PSO}^p$, where $\sigma_h$ and $\sigma_p$ are positive, and $\sigma_h + \sigma_p = 1$. Similarly, an instance of equation (\ref{equation16}) can be expressed as $f_{PSO} = \sigma_h.f_{PSO}^h + \sigma_p.f_{PSO}^p + \sigma_i.f_{PSO}^{i,s}$, with $\sigma_h$, $\sigma_p$, and $\sigma_i$ are positive, and $\sigma_h + \sigma_p + \sigma_i =1$.

For Hungarian Assignment, the cost matrix is defined as in equation (\ref{equation21}), where $det_j^{conf}$  is the detection confidence from the object detector, $X_{i,t-1}^{pen}$ is derived from the tracking history, as described in Section II.B (dependent on the GenTrack variant), and $C_m^{X_{i,t}^s,det_j}$ represents the motion consistency cost, computed as: 

\begin{equation}
\label{equation30}
    C_m^{X_{i,t}^s,det_j} = C_{IoU}^{X_{i,t}^s,det_j}.C_d^{X_{i,t}^s,det_j}
\end{equation}

Here, $det_j = (u_{det}, v_{det}, w_{det}, h_{det})$, and $det_j \in \{D_t\}$. The IoU-based cost is defined as $C_{IoU}^{X_{i,t}^s,det_j} = 1 - IoU_{det_j}^{X_{i,t}^s}$, where $IoU_{det_j}^{X_{i,t}^s}$ measures the intersection of union between $X_{i,t}^s$ and $det_j$. The distance cost is given by $C_d^{X_{i,t}^s,det_j} = \frac{min(|X_{i,t}^{s} - det_j|, d_{o,d})}{d_{o,d}}$, where $d_{o,d}$ is the maximum allowed distance between $X_{i,t}^s$ and $det_j$, obtained by using $(w_i, h_i)$ and $(w_{det}, h_{det})$.

It is noted that image features are consistently available for PSO fitness measure, as both particle states and previous optima are maintained. Besides, IoU is excluded from the PSO fitness function to promote exploration, as IoU biases particles toward their comparison states. In data association, detection confidence and track penalties are integrated into the matching cost, enabling a systematic and efficient matching process. As a result, sub-matchings for unmatched detections and weak tracks, as used in some other trackers, is unnecessary.

Finally, the state update of weak tracks in GenTrack Super as in the first line of equation (\ref{equation25}), which is initially updated based on its global best particle, followed by an adjustment using the states of its neighbours, as:

\begin{equation}
\label{equation31}
\begin{cases}
    X_{t [1:2]}^w = X_{t [1:2]}^w + \sigma_s.\langle \Delta_{pos} \bullet \hat{V}_{t[1:2]}^w \rangle . \hat{V}_{t[1:2]}^w  \\
    X_{t[3:4]}^w = X_{t[3:4]}^w + \sigma_s.(\overline{X}_{j,t[3:4]} - X_{t[3:4]}^w)\\
    \hat{V}_{t[1:2]}^w = \frac{V_{t[1:2]}^w}{|V_{t[1:2]}^w|} |  |V_{t[1:2]}^w| > 0 \\
    \Delta_{pos} = \overline{X}_{j,t[1:2]} - X_{t[1:2]}^w \\
    \overline{X}_{j,t} = \frac{1}{N}.\sum_{j=1}^N X_{j,t}
\end{cases}
\end{equation}

Equation (\ref{equation31}) applies to the object state as defined in Section II.C.1. And $\sigma_s$ is a small positive value.

\section{\textbf{Case Studies}}

The proposed method is validated across various scenarios using both human and cow datasets, demonstrating its current object-agnostic design without reliance on category-specific features. A source-code reference implementation with minimal dependencies is provided, featuring three variants: GenTrack Simple, GenTrack Strengthen, and GenTrack Super. As presented, the source-code reference implementation also offers different approaches for particle generation and post-PSO resampling, allowing for adaptable redevelopment and comparative analysis. Additionally, most parameters are adaptive, while a few are user-tunable with minimal impact on performance. All fitness and cost metrics are normalized to [0, 1] to support systematic analysis. Furthermore, comparative evaluations are conducted against state-of-the-art trackers, including SORT \cite{ref7}, DeepSORT \cite{ref8}, ByteTrack \cite{ref9}, BoT-SORT \cite{ref10}, OC-SORT \cite{ref11}, SMILEtrack \cite{ref13}, and ConfTrack \cite{ref14}, using the same detection model to ensure fair comparisons. Implementations of these trackers are also integrated into the source code, give more favourable conditions to re-conduct comparisons. All experiments are conducted on a single PC configured with an Intel Core i9-14900KF processor, an NVIDIA GeForce RTX 4090 GPU with 24GB of VRAM and 64GB of RAM. GenTrack variants are executed on the CPU (Central Processing Unit), while object detection is performed on the GPU (Graphics Processing Unit).

All trackers were implemented and evaluated across two scenarios, with the same parameters used for each tracker. The performance of the trackers is assessed using the metrics defined in \cite{ref46, ref47}. Higher values of \textbf{ATA} (Average Tracking Accuracy), \textbf{IDF1}, \textbf{HOTA} (Higher Order Tracking Accuracy), and \textbf{MOTA} (Multiple Object Tracking Accuracy) indicate better performance. Conversely, lower values of \textbf{IDSW} (ID switches) reflects better performance. Here, the values of \textbf{ATA}, \textbf{IDF1}, \textbf{HOTA}, and \textbf{MOTA} lie within the range [0, 100].

\subsection{Human Tracking}

The first scenario utilizes the widely recognized MOT17 dataset \cite{ref48}, a benchmark for single-camera multi-object tracking using pinhole camera footage. Given the emphasis on static cameras, the tracking performance is evaluated using sequence MOT17-04, containing 1050 frames at the resolution of 1920x1080, recorded at 30 frames per second (FPS). The objective of this scenario is to track all humans in the scene, although the MOT17 dataset is primarily designed for pedestrian detection and tracking. Consequently, the human detection employed is relatively weak and noisy, which offers a challenging yet fair setting for evaluating tracker performance. To align with the goal of this scenario, the ground truth was modified accordingly and is provided alongside the source code for efficient re-evaluation. Both detection results and ground truth were held constant across all evaluations of trackers to ensure fairness. The revised MOT17-04 sequence retains its original name.

\begin{figure}[!t]
\centering
\includegraphics[width=3.4in]{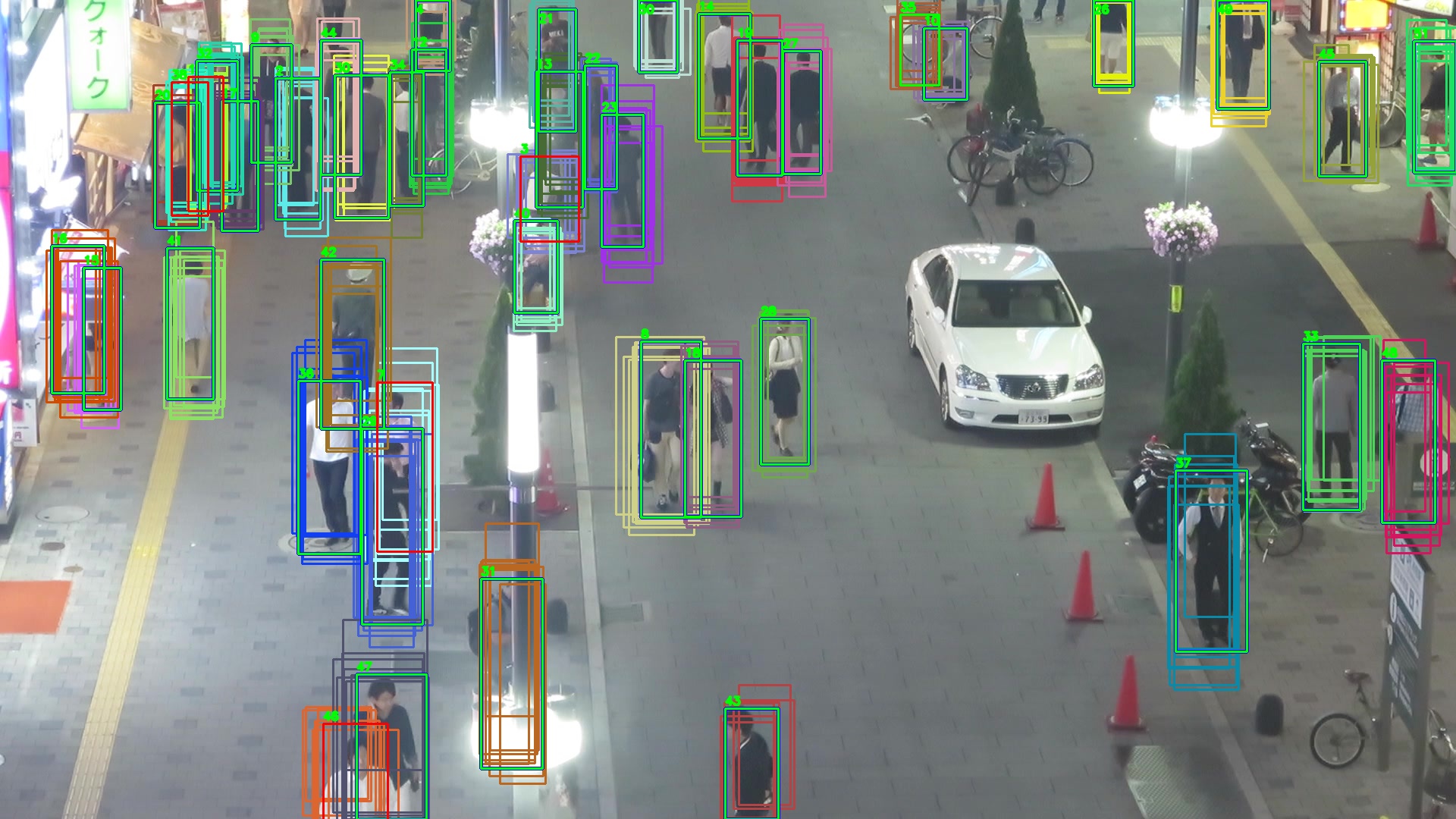}
\caption{An illustration of visual human tracking, with particle visualization.}
\label{fig_1}
\end{figure}

\begin{figure}[!t]
\centering
\includegraphics[width=3.4in]{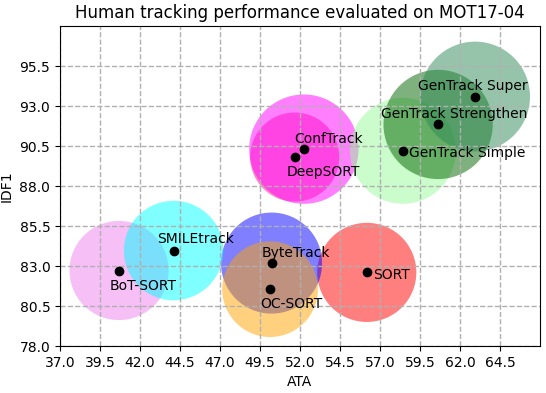}
\caption{ATA-IDF1-HOTA comparisons of trackers on human tracking.}
\label{fig_2}
\end{figure}

An illustration of human tracking is provided in Fig. \ref{fig_1}, with results summarized in Table \ref{tab:table1} and Fig. \ref{fig_2}. A visualization of tracking output of GenTrack conducted on MOT17-04 can be found in Seq. 1, Table \ref{tab:table3}, Section III. C. 

\begin{table}[!t]
    \caption{Human Tracking Performance Evaluated on the MOT17-04 Sequence \cite{ref48}\label{tab:table1}}
    \centering
    \scriptsize
    \setlength{\tabcolsep}{5pt}
    \renewcommand{\arraystretch}{1.1}
    \begin{tabular}{|l||c||c||c||c||c|}

    \hline
    {\textbf{Tracker}} & {\textbf{ATA}} & {\textbf{IDF1}} & {\textbf{HOTA}} & {\textbf{MOTA}} & {\textbf{IDSW}}\\
    \hline
    {SORT \cite{ref7}} & {56.174} & {82.600} & {61.962} & {69.860} & {103} \\
    \hline
    {DeepSORT \cite{ref8}} & {51.668} & {89.808} & {55.701} & {75.574} & {11}\\
    \hline
    {ByteTrack \cite{ref9}} & {50.224} & {83.181} & {63.169} & {74.394} & {85}\\
    \hline
    {BoT-SORT \cite{ref10}} & {40.690} & {82.718} & {61.970} & {71.417} & {316}\\
    \hline
    {OC-SORT \cite{ref11}} & {50.128} & {81.551} & {59.802} & {69.628} & {280}\\
    \hline
    {SMILEtrack \cite{ref13}} & {44.101} & {83.965} & {62.172} & {71.484} & {278}\\
    \hline
    {ConfTrack \cite{ref14}} & {52.241} & {90.302} & {68.345} & {82.778} & {58}\\
    \hline
    {GenTrack Simple} & {58.432} & {90.200} & {66.226} & {80.416} & {12}\\
    \hline
    {GenTrack Strengthen} & {60.632} & {91.847} & {68.290} & {83.379} & {10}\\
    \hline
    {GenTrack Super} & {\textbf{62.946}} & {\textbf{93.593}} & {\textbf{68.601}} & {\textbf{85.111}} & {\textbf{4}}\\
    \hline

\end{tabular}
\end{table}

\subsection{Cow Tracking}

Fisheye cameras, due to their wide field of view (FOV), have gained traction across various applications \cite{ref49,ref50,ref51}. As fisheye cameras have emerged as promising alternatives in a spectrum of real-life applications, this scenario employs the MooTrack360 \cite{ref52}, designed for tracking dairy cows – an emerging application in livestock monitoring. Moreover, cow tracking provides a valuable benchmark for evaluating tracking performance due to the inherently unpredictable nature of animal movement and the absence of a reliable motion model. The tracking performance is evaluated using 1-hour video sequence (corresponding to 53920 frames) at the resolution of 2944x2944 pixels, with the video speed is 15 FPS. Frames are extracted and resized to 1472x1472 pixels before being processed by the detector. Accordingly, ground truth bounding boxes are scaled to 1472x1472 pixel coordinates to ensure consistency in performance metric computation. 

\begin{figure}[!t]
\centering
\includegraphics[width=3.2in]{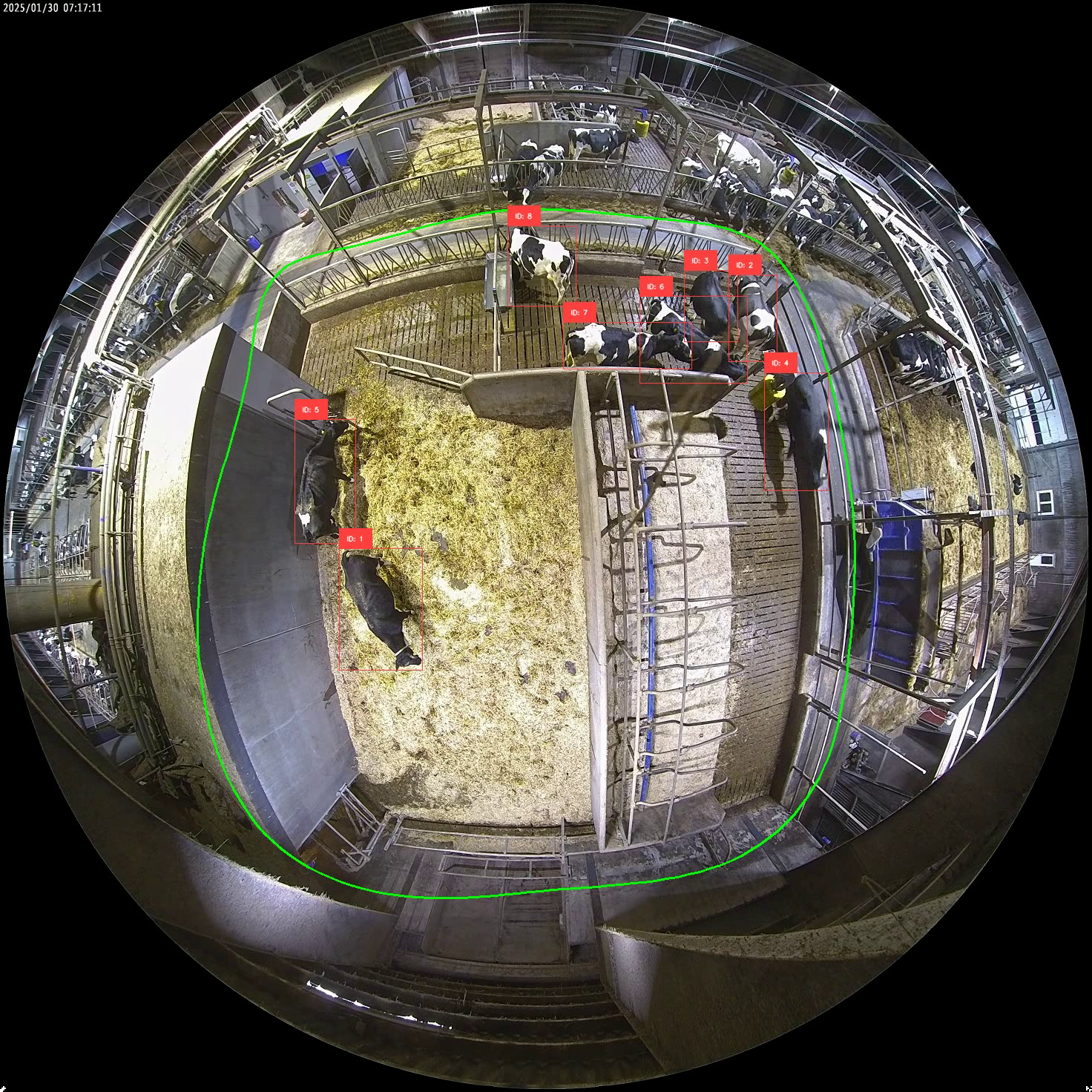}
\caption{An illustration of visual cow tracking, without particle visualization.}
\label{fig_3}
\end{figure}

\begin{figure}[!t]
\centering
\includegraphics[width=3.3in]{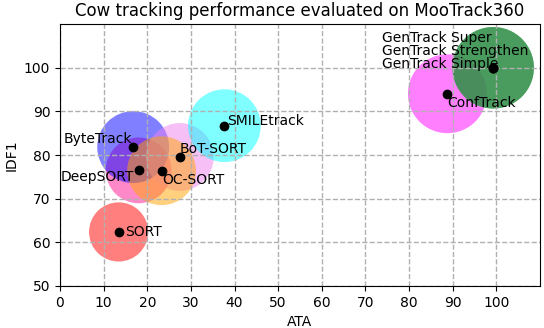}
\caption{ATA-IDF1-HOTA comparisons of trackers on cow tracking.}
\label{fig_4}
\end{figure}

An illustration of cow tracking and monitoring can be found in Fig. \ref{fig_3}, with results summarized in Table \ref{tab:table2} and Fig. \ref{fig_4}. The tracking output visualization of GenTrack conducted on MooTrack360 is available in Seq. 2, Table \ref{tab:table3}, Section III. C. 

\begin{table}[!t]
    \caption{Cow Tracking Performance Evaluated on the MooTrack360 1-hour Video Sequence \cite{ref52}\label{tab:table2}}
    \centering
    \scriptsize
    \setlength{\tabcolsep}{5pt}
    \renewcommand{\arraystretch}{1.1}
    \begin{tabular}{|l||c||c||c||c||c|}

    \hline
    {\textbf{Tracker}} & {\textbf{ATA}} & {\textbf{IDF1}} & {\textbf{HOTA}} & {\textbf{MOTA}} & {\textbf{IDSW}}\\
    \hline
    {SORT \cite{ref7}}          & {13.435} & {62.370} & {56.485} & {99.241} & {137}\\
    \hline
    {DeepSORT \cite{ref8}}      & {18.016} & {76.472} & {62.778} & {96.924} & {35}\\
    \hline
    {ByteTrack \cite{ref9}}     & {16.778} & {81.809} & {68.675} & {99.468} & {20}\\
    \hline
    {BoT-SORT \cite{ref10}}     & {27.456} & {79.543} & {64.781} & {98.823} & {21}\\
    \hline
    {OC-SORT \cite{ref11}}      & {23.280} & {76.399} & {65.742} & {98.068} & {94}\\
    \hline
    {SMILEtrack \cite{ref13}}   & {37.678} & {86.718} & {69.468} & {98.842} & {11}\\
    \hline
    {ConfTrack \cite{ref14}}    & {88.777} & {94.037} & {75.322} & {99.982} & {1}\\
    \hline
    {GenTrack Simple}           & {\textbf{99.266}} & {\textbf{99.992}} & {\textbf{77.832}} & {\textbf{99.984}} & {\textbf{0}}\\
    \hline
    {GenTrack Strengthen}       & {\textbf{99.266}} & {\textbf{99.992}} & {\textbf{77.832}} & {\textbf{99.984}} & {\textbf{0}}\\
    \hline
    {GenTrack Super}            & {99.264} & {\textbf{99.992}} & {77.830} & {\textbf{99.984}} & {\textbf{0}}\\
    \hline
    
    \end{tabular}
\end{table}

\subsection{GenTrack Visualization}

Table \ref{tab:table3} presents a visualization of GenTrack’s experimental results on both MOT17-04 and MooTrack360, with “Seq” denoting the event sequence and “Snapshot” indicating the video action start time. To ensure clear and rapid visual interpretation, human tracking video is displayed at half speed and cow tracking video at ten times the speed relative to the original inputs. The actual latency of GenTrack is detailed in the Section IV. 

In Seq. 1 (human tracking), the visualization serves primarily for debugging: cyan bounding boxes indicate strong tracks with matched detections, red bounding boxes indicate weak tracks, green bounding boxes indicate recovered weak tracks with high fitness scores, blue bounding boxes indicate unmatched detections. For cow tracking in Seq. 2, the visualization targets application use, with bounding box colors denoting object classes – red for “cow standing” and purple for “cow lying”. Particle visualization is also supported in the source-code and can be easily implemented. 

\begin{table}[!t]
    \caption{Visualization of the Experimental Results from GenTrack.\label{tab:table3}}
    \centering
    \scriptsize
    \setlength{\tabcolsep}{5pt}
    \renewcommand{\arraystretch}{1.1}
    \begin{tabular}{|c||p{3.6cm}||p{2.8cm}|}

    \hline
    {Seq} & {Event} & {Snapshot (mm:ss)}\\
    \hline
    {1} & {GenTrack Case Studies: Human tracking based on MOT17-04 sequence.} & {\href{https://www.youtube.com/watch?v=b9CzP4bENno\&t=4s}{\url{https://www.youtube.com/watch?v=b9CzP4bENno\&t=4s}} (00:04)} \\
    \hline
    {2} & {GenTrack Case Studies: Cow tracking by using MooTrack360 1-hour video sequence.} & {\href{https://www.youtube.com/watch?v=b9CzP4bENno\&t=77s}{\url{https://www.youtube.com/watch?v=b9CzP4bENno\&t=77s}} (01:17)}\\
    \hline
    \end{tabular}
\end{table}

\section{\textbf{Discussions}}

Tracker performance is evaluated in short-term using the human tracking scenario with 1050 frames, and in the long-term using the cow tracking scenario with 53920 frames. The experimental results from case studies demonstrate that the proposed method, GenTrack, consistently outperforms the compared state-of-the-art trackers in both human and cow tracking tasks, as evidenced by Tables I-II, Fig. 2, and Fig. 4. Particularly, the proposed GenTrack handles the cow tracking scenario with ease, achieving success rates of 100 percent, without ID switches. In comparison, methods SORT \cite{ref7}, DeepSORT \cite{ref8}, ByteTrack \cite{ref9}, BoT-SORT \cite{ref10}, OC-SORT \cite{ref11}, SMILEtrack \cite{ref13}, and ConfTrack \cite{ref14} require significant tuning time but still fail to achieve a 100 percent success rate, in which ConfTrack \cite{ref14} perform best, with only one ID switch. In human tracking scenario, the trackers are implemented with the same parameters used in cow tracking scenario. Among them, GenTrack Super achieves the best performance, with just 4 ID switches. Further performance metrics are detailed in Tables I-II, Fig. 2, and Fig. 4.

Insights from implementation reveal that GenTrack exhibits performance largely insensitive to parameter settings. The source code supports particle generation from both the previous optimal state and prior particles; however, the former is recommended due to the absence of a motion model. Using a random motion model for all prior particles may cause divergence from the convergence point, which requires additional post-processing. Moreover, post-PSO resampling options are available to better align particles with the distribution modes. Global and discard resampling yield comparable results, but when using a small number of particles and discarding low fitness particles, discard thresholds must be chosen carefully to avoid eliminating informative particles. Finally, noise bounds are set by the bounding box dimensions per axis, while inter-state motion is bounded by the sum of the diagonals of the two bounding boxes.

The latency of GenTrack variants is directly influenced by the number of particles, with higher counts resulting in increased latency. However, unlike conventional particle-based methods, GenTrack maintains robust performance with a small particle set – 8 particles for GenTrack Simple, and 6 particles for both GenTrack Strengthen and GenTrack Super to achieve performance as demonstrated in Tables I-II, Fig. 2 and Fig. 4. Computational analysis confirms GenTrack’s suitability for real-time tracking, with CPU-based latencies of 3.67(ms), 6.35 (ms), and 6.74 (ms) for GenTrack Simple, GenTrack Strengthen, and GenTrack Super, respectively, in the cow tracking scenario. In the more complex human tracking scenario, latencies rise to 56 (ms), 64 (ms), and 92 (ms), respectively. These latencies were evaluated without GPU acceleration, and the source code is in scalar styles. The latency can be further reduced by vectorizations of the source code, and will be provided in the next version.

This paper introduced the GenTrack philosophy and its application to visual multi-object tracking, supporting a systematic tracking pipeline. Tracking performance can be improved through specific considerations:

\begin{itemize}[leftmargin=*]
\item{\textit{Appearance similarity}: This paper measures appearance similarity for visual multi-object tracking using cosine similarity between standard HoG vectors extracted from bounding boxes. The HoG features are computed in a straightforward manner without object-specific processing. Enhancing appearance similarity measures tailored to specific targets and applications could improve the performance of GenTrack-based tracking systems. Furthermore, it may be better to compare appearance similarity within the object regions instead of the entire bounding boxes to ignore background effects.}
\item{\textit{Noise-adaptable sampling}: Although low-fitness particles can be discarded or replaced by global bests post-PSO, a noise-dependent particle initialization strategy may be introduced to reduce redundancy and improve performance in noisy environments – allocating more particles to regions with higher noise. For instance, the Metropolis-Hastings algorithm may be employed for proposal generation, with PSO handling the exploration.}
\item{\textit{Improved motion model}: The current tracking system employs a random motion model, which can be enhanced by incorporating path-based dominant-set clustering. This integration would improve target motion modelling and better address occlusions and ID switches, particularly during prolonged occlusions.}
\end{itemize}

Furthermore, several promising research directions related to GenTrack may include the following: 

\begin{itemize}[leftmargin=*]
\item{\textit{Group tracking}: A novel GenTrack variant for group tracking can be developed by integrating a clustering algorithm, such as DBSCAN, to aggregate individuals, alongside group operations, including member adding-removing, group merging-splitting. }
\item{\textit{Multi-cam multi-object tracking}: An extended GenTrack framework can be conducted for multi-cam MOT, applicable both with and without overlaps. Tracking may be conducted separately within each camera, followed by inter-camera track association, or by integrating multi-camera detections into a unified frame for single-camera tracking. In non-overlapping scenarios, targets are treated as undergoing prolonged occlusions when moving between cameras, and their association is achieved via path-based dominant-set clustering.}
\item{\textit{Three-dimensional (3D) multi-object tracking}: Redefinition of object state and observation models to incorporate 3D data, utilizing either 3D object detectors or clustering methods for object detection.}
\end{itemize}

\section{\textbf{Conclusions}}
This paper presents a novel multi-object tracking (MOT) framework, named GenTrack, which restructures existing MOT baselines to facilitate a more systematic tracking pipeline. Several variants of GenTrack are developed and implemented for visual multi-object tracking, with empirical validation conducted on human and cow tracking datasets. Comparative analyses against state-of-the-art trackers demonstrate that the proposed GenTrack framework achieves better performance. Furthermore, it is observed that tracking efficacy can be further enhanced through the careful design of internal and observation models tailored to specific target objects and application contexts. Finally, potential extensions and promising directions for future research pertaining to GenTrack are discussed.

\section*{Acknowledgments}
This work was a part of VelKoTek project, funded by The Ministry of Food, Agriculture and Fisheries of Denmark via the Green Development and Demonstration Program (GUDP).


\end{document}